# Automatic Assessment of Alzheimer's Disease Diagnosis Based on Deep Learning Techniques *

**Alejandro Puente-Castro**[†]
Faculty of Computer Science, CITIC
University of A Coruna
A Coruna, 15007, Spain

**Enrique Fernandez-Blanco**
Faculty of Computer Science, CITIC
University of A Coruna
A Coruna, 15007, Spain

**Alejandro Pazos**
Faculty of Computer Science, CITIC
University of A Coruna
A Coruna, 15007, Spain
Biomedical Research Institute of A Coruña (INIBIC)
University Hospital Complex of A Coruna (CHUAC)
A Coruna, 15006, Spain

**Cristian R. Munteanu**
Faculty of Computer Science, CITIC
University of A Coruna
A Coruna, 15007, Spain
Biomedical Research Institute of A Coruña (INIBIC)
University Hospital Complex of A Coruna (CHUAC)
A Coruna, 15006, Spain

## Abstract

Early detection is crucial to prevent the progression of Alzheimer's disease (AD). Thus, specialists can begin preventive treatment as soon as possible. They demand fast and precise assessment in the diagnosis of AD in the earliest and hardest to detect stages. The main objective of this work is to develop a system that automatically detects the presence of the disease in sagittal magnetic resonance images (MRI), which are not generally used. Sagittal MRIs from ADNI and OASIS data sets were employed. Experiments were conducted using Transfer Learning (TL) techniques in order to achieve more accurate results. There are two main conclusions to be drawn from this work: first, the damages related to AD and its stages can be distinguished in sagittal MRI and, second, the results obtained using DL models with sagittal MRIs are similar to the state-of-the-art, which uses the horizontal-plane MRI. Although sagittal-plane MRIs are not commonly used, this work proved that they were, at least, as effective as MRI from other planes at identifying AD in early stages. This could pave the way for further research. Finally, one should bear in mind that in certain fields, obtaining the examples for a data set can be very expensive. This study proved that DL models could be built in these fields, whereas TL is an essential tool for completing the task with fewer examples.

## 1 Introduction

One of the main consequences of the progressive aging of the population is a higher occurrence of age-related neurodegenerative diseases. Among these diseases, Alzheimer's disease (AD) stands out with a prevalence of 5.5% in Europe in 2016 [1] and 10% in the United States of America in 2019 [2].

The main challenge faced by the AD researchers today is to perform a *pre mortem* diagnosis that leaves no room for doubt. Many things are unknown about this disease today besides the symptoms of mood swings and some obvious changes in the morphology of the cerebral cortex [3], which are not related to a normal aging process [4]. Although the mechanisms of AD have not yet been unveiled, most experts agree that the disease seems to be the result of a combination of genetic and environmental factors. For example, some studies have related it to pathogens causing periodontitis [5] or to herpes simplex virus type 1 [6]. It is believed that it is an age-related but not an age-dependent disease [7] and that there is sexual dimorphism in the presence of the disease [8]. Even today, the disease is the subject of in-depth studies due to its impact on society and the lack of agreement on its origin.

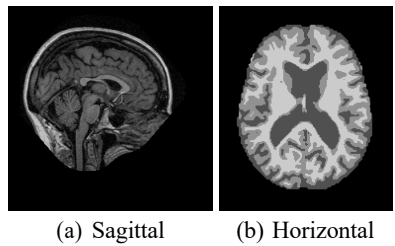

(a) Sagittal    (b) Horizontal

Figure 1: Examples of the different types of MRI according to the acquisition plan. They display different changes in the morphology of the brain organ depending on the type.

Nowadays, very few things are clear about this neurodegenerative disease but it has no cure apart from a series of palliative treatments to slow down its progression. Therefore, early detection of the disease is a key element for improving the quality of life of affected patients and their families. To make this diagnosis, cognitive, psychological or clinical tests are used. In clinical trials, among the most used, are the brain images obtained by magnetic resonance imaging (MRI). This is due to the fact that these images hold a high relationship with the brain topology and they show the alterations in the brain morphology. Being able to see the alterations in brain morphology is the main reason why MRI images are used. In the MRI images, the regions with the cells affected by the degeneration of the disease take very low-intensity values, so they appear darker with respect to healthy parts (Figure 1).

Focusing on techniques to support the diagnostics of AD based on images, most of them are related to Computer Vision [9], although Machine Learning [10] techniques have recently been introduced. For example, Stonnington et al. [11] used regression models with likelihood functions for screening and tracking the disease, whereas Li et al. [12] employed Vector Support Machines (SVM) for diagnostic support.

Recently, advances have been introduced in Machine Learning such as Deep Learning (DL) techniques [13]. Liu et al. [14] proposed the use, in their medical application, of autoencoders with Softmax output layers to avoid bottleneck during diagnostic support. The utilization of convolutional networks is widespread, as in the works conducted by Hosseini et al. [15], because they are the most adapted type for the treatment of images and signals. Zheng et al. [16] proposed the use of restricted Boltzmann machines as a preliminary stage to a classic Machine Learning classifier to improve its results.

Other proposals, such as that of Suk et al. [17], used volumes of gray matter (GM volumes) instead of MRI images for the diagnosis of AD, and thus were able to obtain information from a large part of the brain despite the computational cost. In this work and in some other proposals, they extracted information from various medical imaging techniques such as combined information from the horizontal MRI images and their corresponding positron-emission tomography (PET) images.

The use of DL was also combined with MRI images for other applications in the field of neuroscience, not just the diagnosis of AD as in Suk's work [18], in which a method was proposed for the identification of activity changes in brain regions during rest.

All of these examples employed their own complex architectures with horizontal MRI images or volumes. In some cases, horizontal MRI images were combined with other types of images such as sagittal MRI images. Unlike the previously described works, the present work proposes the exclusive use of sagittal images along with fine-tuning of general-purpose architectures into images. This would allow experiments to be carried out in a more exhaustive way, because it has information about the brain at different vertical levels, and at a lower computational cost, since there would be no need to process a great amount of images. This would also make possible to extract information from other regions of the brain.

As seen in some examples listed above, DL allows characterizing AD in MRI images through the creation of computational models composed of multiple processing layers. Despite being an approximation of Machine Learning, DL automatically extracts its own information from the input images [19], avoiding the subjectivity of the expert who labels the information, as in a classic Machine Learning model.

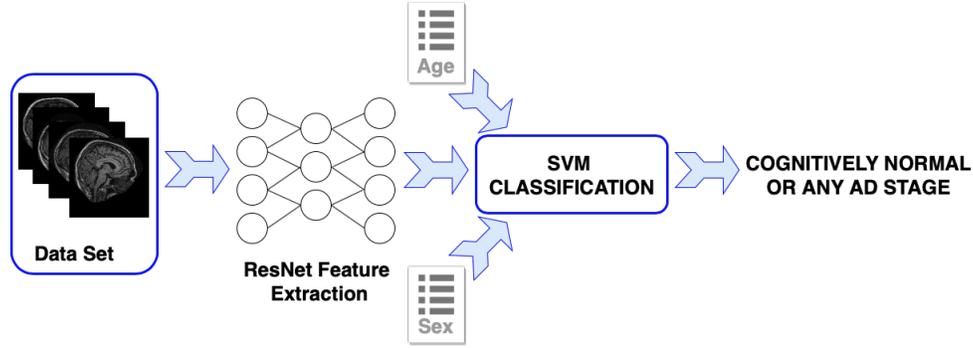

Figure 2: The workflow diagram of the study. Patients' MRI scans were fed to a ResNet ANN [26] in order to extract new features vectors and sex and age are concatenated to them. These vectors are separated into training data and test data. Test data is used for training an SVM model [27]. Test data is used for evaluating trained SVM model goodness in order to improve it.

It is important to consider the correct use of DL in the solution of the problem. In the ADNI [20] data set, Basaia et al. [21] employed Data Augmentation to overcome the scarce number of data. This could be risky because the transformations of deformation and crop may not represent real cases or errors and they can be multiplied in case samples were erroneously labeled. In the OASIS [22] data set, the use of a small and efficient network by Islam et al. [23] stands out. Despite employing Data Augmentation, the utilization of the Hold-Out validation strategy in this work may not be a very comprehensive measurement of the model's goodness of fit due to the large imbalance present between the classes of the data set and its small size. Removing 20% of the training data involves the risk of eliminating underrepresented classes from the model training stage. Usually, other works obtained their performance metrics by contrasting the values of each class against the other. Therefore, they treated problems as if they were independent binary problems. For example, Yue et al. [24] addressed the problem as three separate binary problems, one for each class. The results of the prediction were obtained for each class in the presence of each of the remaining classes. Therefore, biased values were obtained, since the values of these binary comparisons were not achieved in the presence of the noise from the other classes.

In conclusion, the objective of this work is the development of a DL application that assesses the diagnosis of AD in sagittal MRI images. Therefore, it is expected to achieve, as main results: a valid model developed with DL to perform the diagnosis with sagittal MRIs; to prove that sagittal images can be used in order to identify AD, without the help of any other kind of acquisition plane; and to prove that Transfer Learning is crucial for training models with few data.

## 2 Materials and Methods

### 2.1 Workflow

The schema shown in Figure 2 illustrates the main workflow of this study. It is similar to the classical pipeline to process any kind of signal but adapted to include Transfer Learning (TL) [25]. Patients' MRI scans were fed to a ResNet artificial neural network (ANN) [26] in order to extract new feature vectors and patients' sex and age are concatenated to them. The obtained vectors contained 100,352 numerical values extracted from each MRI scan. The extracted measures represent those different measures that ResNet model considers relevant for learning. It is easier for the ANN to learn some important features from an image instead of learning the complete image and the spatial relationships between pixels. The full set of vectors was divided into training and testing data. The training data were used for training an SVM model [27] with Radial Basis Function (RBF) kernel. The testing data were employed for evaluating trained SVM model goodness in order to improve it. The main reason to use TL is the small size that MRI data sets have. In this way, the chosen model was trained for another task with a huge amount of images, such as different types of medical images. Therefore, fewer weights were required to be tuned, which implies that less new data were needed to adapt the model to the current problem. Consequently, using ANN trained in the previous tasks required less amount of data and, collaterally, the experimentation was faster.

Going along the workflow, first, the information about the patients was loaded, that is their age, sex, and the patients' MRI scans scaled to 224x224 pixels. This last requirement was set by the trained ANN used in this work which is known as ResNet [26]. Second, all MRI scans were fed to the ResNet ANN without the last three layers in order to automatically extract feature vectors. That feature vector was concatenated to their respective sex and age values. Finally, in the last step, the data set was split into training and testing data. Training data were used to develop an SVM model [27] while testing data were used to measure the behavior of the model with previously unseen data.

## 2.2 Data Sets

This work made use of two sets of images known as OASIS and ADNI. Both sets were properly labeled collections of MRI images, which are two of the most common in the state-of-the-art.

### 2.2.1 OASIS

The OASIS data set [22] is open and presents two collections: OASIS-Cross-sectional, which contains images of sagittal MRI, and OASIS-Longitudinal, which is formed by longitudinal slices. Both collections correspond to the acquisition of 256x256 pixel images from different patients. The objective of this work was the OASIS-Cross-sectional collection because it led to the resolution of the problem with 436 sagittal images. Out of these, 2 were moderate AD cases, 28 are mild dementia, 70 very mild dementia and 316 were cognitively normal. This implies a strong imbalance in favor of cognitively normal cases. The images were obtained from 168 male patients and 268 female patients aged between 18 and 98 years.

### 2.2.2 ADNI

The ADNI data set [20] contains several collections of MRI images. The MP-RAGE REPEAT collection was used in this work. This collection consists of 1743 volumes from which the cut 88 of the sagittal plane was extracted, corresponding with a 256x256 pixel image that is the central cut. These images were divided into 297 AD cases, 921 mild cognitive dementia, and 525 cognitively normal cases. This implies an imbalance in favor of the AD class. All images were obtained from 1055 male patients and 688 female patients aged between 55 and 94 years.

## 2.3 Proposed Method

The proposed model is shown in Figure 3. This model employed the first 47 layers of a ResNet ANN [26] trained with ImageNet data set to extract the features. ResNet or Residual Neural Network is an ANN based on Residual blocks for avoiding gradient problems that deep ANNs have. Without gradient problems, it is possible to train deeper ANNs.

The output of the last layer of the ResNet was concatenated to the patient's sex and age and then it was classified by a Support Vector Machine (SVM) [27] specifically trained for the problem. The reason why this model composition strategy was chosen is TL [25] technique. Therefore, the speed of experimentation can be accelerated by adapting successful models to related problems.

## 2.4 Performance Measures

In order to show the goodness-of-fit of the proposed model, a bunch of metrics defined with confusion matrix were used (see Table 1). More specifically, in this work Accuracy (1), Precision (2), Recall (3), Specificity (4) and $F_1$ score (5) were chosen because they are the most commonly used in the bioinformatics literature in order to increase the possibilities to compare the results like in the studies [28] and [29].

Table 1: Example of confusion matrix for binary problems.

|        |       | Predicted            |                      |
|--------|-------|----------------------|----------------------|
|        |       | True                 | False                |
| Actual | True  | True Positives (TP)  | False Negatives (FN) |
|        | False | False Positives (FP) | True Negatives (TN)  |

$$Accuracy = \frac{TP + TN}{TP + TN + FP + FN} \qquad (1)$$

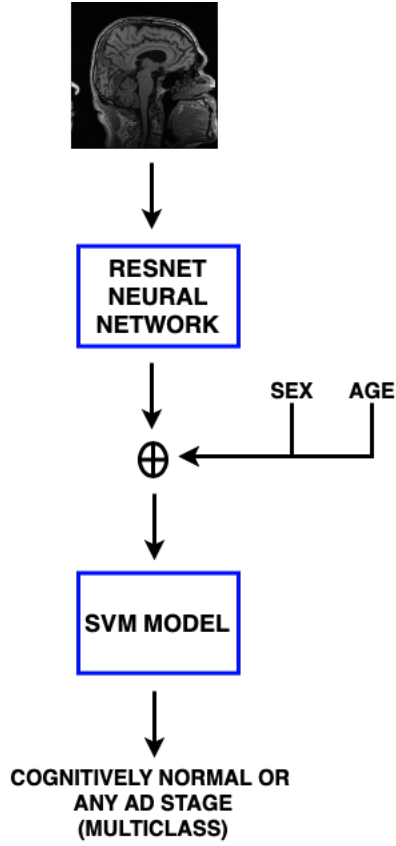

Figure 3: Proposed model schema. Once ResNet ANN [26] extracts feature vectors from MRI scans, sex and age are concatenated in order to add more features. These extended features vectors are fed to the SVM model [27], which determines wheter there is any stage of AD or the patient is cognitively normal.

$$Precision = \frac{TP}{TP + FP} \quad (2)$$

$$Recall = \frac{TP}{TP + FN} \quad (3)$$

$$Specificity = \frac{TN}{TN + FP} \quad (4)$$

$$F_1 = 2 \cdot \frac{Precision \cdot Recall}{Precision + Recall} \quad (5)$$

## 3 Results

### 3.1 Imbalance

Due to the imbalance of the data sets, their empirical searches for each class were performed. These searches allowed for preventing overfitting of the model constructed for each data set and memorizing the most represented class, instead of the least represented one. In these searches, the way the predictions for each class changed was controlled as the weights were modified, especially the interest classes. Therefore, the weight adjustment was balanced to have a value proportional to the number of elements in each class (6). That is, the least represented class will be given the greatest

Table 2: Best results with the OASIS data set.

| | Class | Accuracy | Precision | Recall | Specificity | $F_1$ |
|---|---|---|---|---|---|---|
| [23] (Horizontal MRI) | Cognitively Normal | - | 99.00% | 99.00% | - | 99.00% |
| | Very Mild Dementia | - | 75.00% | 50.00% | - | 60.00% |
| | Mild Dementia | - | 63.00% | 71.00% | - | 67.00% |
| | Moderate AD | - | 33.00% | 50.00% | - | 40.00% |
| | Average | - | 67.50% | 67.50% | - | 66.50% |
| Proposed Model, without considering sex and age (Sagittal MRI) | Cognitively Normal | 79.36% | 89.94% | 82.44% | 69.00% | 86.02% |
| | Very Mild Dementia | 74.31% | 33.06% | 58.57% | 77.32% | 42.27% |
| | Mild Dementia | 92.66% | 0.00% | 0.00% | 99.02% | 0.00% |
| | Moderate AD | 99.54% | 0.00% | 0.00% | 100.00% | 0.00% |
| | Average | 86.47% | 30.75% | 35.25% | 86.34% | 32.07% |
| Proposed Model (Sagittal MRI) | Cognitively Normal | 80.05% | 92.54% | 81.25% | 78.00% | 86.53% |
| | Very Mild Dementia | 75.00% | 35.77% | 70.00% | 75.96% | 47.34% |
| | Mild Dementia | 92.66% | 0.00% | 0.00% | 99.02% | 0.00% |
| | Moderate AD | 99.54% | 0.00% | 0.00% | 100.00% | 0.00% |
| | Average | 86.81% | 32.08% | 37.81% | 88.25% | 33.47% |

Comparison of the sagittal plane (436 cases) against the horizontal plane (436 cases + data augmentation cases). Model compared against the proposed model may be learning non-real cases because of the artificially created cases.

weight or priority. In this way, when multiplying this weight by the number of elements in each class, the same value will always be given.

$$weight(x) = \frac{\text{training examples}}{\text{classes} \times \text{training examples class } x} \quad (6)$$

For example, taking a data set in which there are 8 elements of class 1 and 2 of class 0, class 0 would have the weight **2.5** and class 1 would have the weight 0.625. Hence, multiplying the weight of each class by the number of elements in that same class, the result would be 5. Which, multiplied by the number of classes, would equal the number of elements in the data set (balanced weights).

### 3.2 OASIS

For the OASIS data set, a Leave-One-Out evaluation [30] was performed with 1% of the training data used as the validation set. During the training Early Stopping [31] was performed in order to prevent overfitting. Hyperparameter optimization was carried out by random search.

As shown in Table 2, the results, with and without considering sex and age, for the most advanced stages of AD were not detected (recall equals 0%). This is due to the few cases contained in the data set for these stages (2 cases for moderate AD and 28 for mild dementia) out of the 416 available cases. Performing Data Augmentation, like in [23], would increase these cases and solve imbalance, but the new data remained dependent on the original data. Therefore, the model may not be learning new phenotypic manifestations that represent those stages and keep learning the same cases. In spite of this, the proposed model trained with this data set showed a great capacity to diagnose AD at the earliest stage. In [23], the authors proposed a model which was not as good at the earliest stage, since they detected fewer positive cases (recall equals 50%). As this is a medical problem, in which it is important to find positive cases as soon as possible, Data Augmentation was employed to learn how to address the diagnosis in later stages in a more precise way, at the cost of detecting fewer cases at the earliest stage.

Results when considering sex and age were compared with the results obtained from pure MRI images, showing a small improvement, especially, for the first two classes. The computational cost of obtaining the sex and age variables is low and does not contain noise, as they are demographic variables. Therefore, more precise results, with almost no computational cost or noise, may be obtained. Both when considering and not considering the sex and age variables, precision and recall values of 0% were obtained for the most advanced stages. This is due to the fact that there were only 2 examples of one class and 28 of the other; these are very few cases compared to the 386 of the remaining two. Because of not using Data Augmentation, these results cannot be improved. The main reason for not using this technique is that there is no guarantee that the synthetic data would be cataloged in the same way as the original data. Moreover, these new data can distort the learning process and make the model process the cases as real even though they are impossible to find in nature.

Table 3: Best results with the ADNI data set.

| | Class | Accuracy | Precision | Recall | Specificity | $F_1$ |
|---|---|---|---|---|---|---|
| [15] (Horizontal MRI) | Cognitively Normal | - | 100.00% | 100.00% | - | 100.00% |
| | Mild Cognitive Dementia | - | 60.00% | 80.00% | - | 69.00% |
| | AD | - | 70.00% | 47.00% | - | 56.00% |
| | Average | - | 76.67% | 75.67% | - | 75.00% |
| Proposed Model without considering sex and age (Sagittal MRI) | Cognitively Normal | 78.25% | 64.44% | 59.80% | 86.05% | 62.03% |
| | Mild Cognitive Dementia | 71.51% | 69.02% | 84.32% | 56.95% | 75.91% |
| | AD | 86.40% | 73.42% | 31.88% | 97.62% | 44.45% |
| | Average | 78.72% | 68.96% | 58.66% | 80.21% | 60.79% |
| Proposed Model (Sagittal MRI) | Cognitively Normal | 78.36% | 63.69% | 59.60% | 85.97% | 61.58% |
| | Mild Cognitive Dementia | 71.50% | 69.00% | 84.61% | 56.50% | 76.01% |
| | AD | 86.05% | 73.93% | 30.62% | 97.72% | 43.31% |
| | Average | 78.64% | 68.87% | 58.28% | 80.06% | 60.30% |

Comparison of the sagittal plane (1743 cases) against the horizontal plane (210 cases). Model compared against the proposed model needs to learn fewer cases, having better results but being more overfit.

An example of a learning curve based on the accuracy metric can be seen in Figure 4. One can see that the model maintains a high percentage of accuracy for testing as the training examples increase. This shows that the data set covers a wide variety of cases despite its small size.

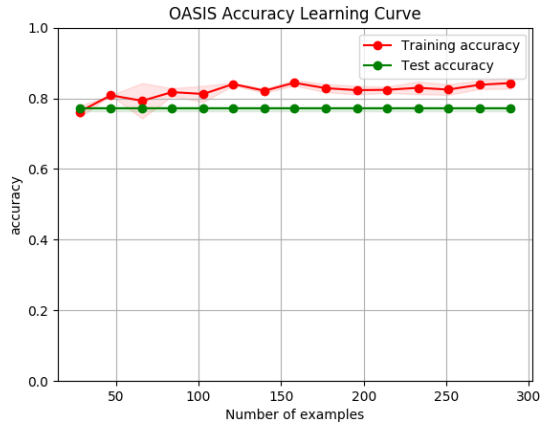

Figure 4: OASIS learning curve with accuracy metric with subsets containing from 10% to 70% of data.

### 3.3 ADNI

For the ADNI data set, 50 repetitions of Hold-Out evaluation [32] were performed with 80% of the entire set used for training, 20% for testing and 1% of the training data used as the validation set. Like the OASIS data set, during the training, Early Stopping [31] was performed in order to prevent overfitting. Hyperparameter optimization was carried out by random search. The obtained results are shown in Table 3.

With this data set, the results were better than in the previous data set. This is due to the smaller imbalance between the stages and the larger number of data. Therefore, there were more data, representing a greater phenotypic variety, for each stage. In spite of this, the AD stage was the least correctly detected (recall equals 30.62%) due to the fact that there were fewer example cases. As in the previous case, this model is satisfactory in addressing the diagnosis at the earliest stages.

[15] presented excellent results, both in the detection of healthy cases and in early stages with very efficient models and without the risks of Data Augmentation. In spite of this, the size of the considered population is very small (210 samples), with people aged around 75 years old for each class, where more men than women are sampled. As shown above, age and sex influence the manifestation of AD, where males being more affected, whereas elderly people have the greatest manifestations, so their results may be biased. Compared to the results of the model proposed in this paper, healthy cases are better detected than other cases. Bearing in mind that the main objective is the diagnosis, it is

preferable that cases be better detected in the presence of AD. There are recent approaches such as [33], but most of their results cannot be compared with those proposed as they performed a pair-wise evaluation of classes. There, the obtained results were dependent on the pairs that were compared, without an overall metric per class. Others, however, used other ADNI data sets that were submitted for a challenge, such as [34]. This data set contains more classes than the usual data set. Therefore, it would not be correct to make a direct comparison with these methods, as the exact equivalence of the classes in the two sets is not known.

The results obtained when considering the demographic variables are worse than not taking them into account, in all cases except for recall in the earliest stage of Alzheimer's. This means that although the detection of the other stages is worse, the stage of interest is detected better. Thus, a healthy case is more likely to be identified with a sick case. As this is a medical care project, this is preferable to the opposite. Therefore, cases should rather be detected as early as possible, so that preventive treatment can be started as soon as possible.

The accuracy-based learning curve for ADNI (Figure 5) shows an evolution of accuracy for tests with a decreasing trend. This means that the data set may not have enough representative cases from each stage for all the possible cases. In contrast to the curve for OASIS, it may be due to having more examples, so there may be more examples that present noise. It may also be due to the fact that ADNI considers fewer stages than OASIS, so there may be overlapping cases that should be considered from different stages. This affects the learning process in the model.

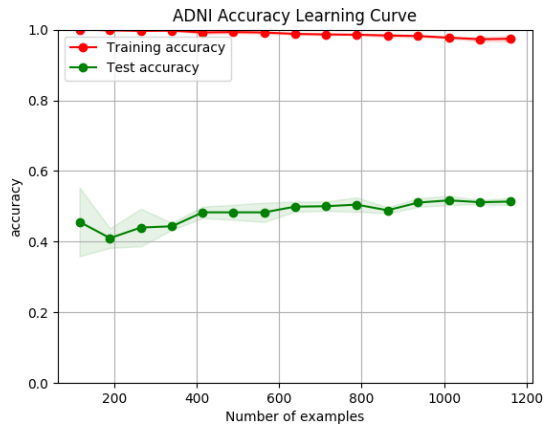

Figure 5: ADNI learning curve with accuracy metric with subsets containing from 10% to 70% of data.

## 4 Discussion

This paper proposes a DL model for the identification of Alzheimer's disease in sagittal MRI images. The extracted data come from two sets of reference data. An initial parameterization was defined in the weights of the classes to overcome the imbalance of data. An evaluation strategy and metrics were established to determine the goodness-of-fit of the proposed model. Therefore, the proposed algorithm presents satisfactory results in both sets with sagittal images.

As the main conclusion, for both data sets, the identification of AD in sagittal MRI images is approachable through DL techniques. These results are comparable to those proposed by horizontal cuts in the literature. Despite the high imbalance of both data sets and the small size of the OASIS set, the proposed model presents satisfactory results for its simplicity compared to those found in the state-of-the-art. This is due to the empirical study of the weights of the classes during training. Therefore, the model knows that the class with the most weight is the one of the greatest interest and has to learn it with the highest priority. The ability to transfer a model from one problem to another opens up the possibility of addressing this problem through TL [25]. Adapting networks that were designed for other problems to a given problem saves time in analysis and network design. By being trained with many different data, ANN does not need many new data to be adapted to the new problem. The fewer the data, the faster the training process is, which speeds up the experimentation. In addition, given that they were trained with different types of images, the greater the capacity for generalization is greater.

From the DL point of view, the overfitting of the data sets and the amount of elements of each class are important aspects. Techniques such as Early Stopping [31] are of great help for this study along with weight adjustment.

According to the related work and the results obtained, it is important to choose a training strategy for the appropriate model. Therefore, the model can cover as many cases as possible with as little training as possible. Adequate model depth is also important, which should be as small as possible. Otherwise, the greater the depth, the more cases the model memorizes. The higher the number of cases the model has learned, the more overfitted it will be. To summarize, the results of the system will be worse when faced with new data.

In both data sets, the proposed model detects better cognitively normal cases and early stages of the disease. This is preferable to the opposite since in the final stage (AD) the presence of the disease is very clear. This model was designed to assist in the early diagnosis of the disease. Therefore, routines and treatments can be started from the earliest stages to slow the progression of AD and prevent it from reaching other later stages.

From the experience of the authors in the field of AD, the sagittal plane also shows typical deformations of the disease. This paves the way for experimentation. New characteristics of AD can be found in other regions. These new characteristics can make diagnosis a more accurate process.

## 5 Conclusion

This study proposes a method of detection of AD using DL techniques and sagittal MRI images. The TL [25] technique was used, using the ANN ResNet [26] feature extractor with the SVM classifier [27]. The model was tested in two sets of reference data, proving its goodness-of-fit by means of previously agreed evaluation strategies and metrics.

The experimental results show that the model is satisfactory compared to previous works with the classical horizontal plane MRI, especially when detecting the initial stages of the AD. These are the most difficult stages to detect, due to the low phenotypic manifestation, and more importantly, to the greater efficacy of the therapy in early stages. This proves that the problem can be approached from the sagittal plane, paving the way for investigation.

TL allows experiments with little data as well as Data Augmentation. Unlike this technique, with the use of TL, there is no risk of generating cases that do not come close to reality or that replicate labeling errors. In addition, using pre-trained models that require fewer data makes processing a task faster, accelerating the experiment design.

In the future, efforts will be made to improve these results by combining information from sagittal MRI volumes. Thereby, sagittal information is available for the entire brain, improving predictions. It is even possible to combine information from other less common planes, such as the frontal plane, along with the sagittal plane, obtaining information from other, less studied, angles. Future work will also focus on improving the model for recording brain regions affected by the disease. Therefore, it may be possible to predict which parts would be affected earlier in the brain and at earlier stages.

### Acknowledgments

The authors would like to thank the support from the CESGA, where many of the tests were run. This work is supported by the "Collaborative Project in Genomic Data Integration (CICLOGEN)" PI17/01826 funded by the Carlos III Health Institute in the context of the Spanish National Plan for Scientific and Technical Research and Innovation 2013–2016 and the European Regional Development Funds (FEDER)—"A way to build Europe." This project was also supported by the General Directorate of Culture, Education and University Management of Xunta de Galicia (Ref. ED431G/01, ED431D 2017/16), the "Galician Network for Colorectal Cancer Research" (Ref. ED431D 2017/23), Competitive Reference Groups (Ref. ED431C 2018/49) and the Spanish Ministry of Economy and Competitiveness via funding of the unique installation BIOCAI (UNLC08-1E-002, UNLC13-13-3503) and the European Regional Development Funds (FEDER). Enrique Fernandez-Blanco would also like to thank NVidia corp., which granted a GPU used in this work for the preliminary tests.

### Supplementary Material

Source code can be found at `https://github.com/TheMVS/DL_AD_mri_sex_age_stages` and a Docker image is available at `https://hub.docker.com/r/themvs/dl_ad_mri_sex_age_stages`.